\documentclass[runningheads]{llncs}
\usepackage{graphicx}

\usepackage{tikz}
\usepackage{comment}
\usepackage{amsmath,amssymb} 
\usepackage{color}

\usepackage[accsupp]{axessibility}  

\usepackage{hyperref}
\usepackage{multirow}
\newcommand\modelname{FairDisCo}
\newcommand\modelwocl{FairDisCo$^{\oslash}$}  

\begin{document}
\pagestyle{headings}
\mainmatter
\def\ECCVSubNumber{110}  

\title{\modelname{}: Fairer AI in Dermatology via Disentanglement Contrastive Learning} 

\titlerunning{\modelname{}: Fairer AI in Dermatology}
%
\author{Siyi Du\inst{1}\index{van Author, First E.} \and
Ben Hers\inst{1} \and
Nourhan Bayasi\inst{1} \and
Ghassan Hamarneh\inst{2} \and  
Rafeef Garbi\inst{1} }

%
\authorrunning{S. Du et al.}
%
\institute{University of British Columbia, Vancouver, British Columbia, CA \and
Simon Fraser University, Burnaby, British Columbia, CA \\
\email{\{siyi,bhers,nourhanb,rafeef\}@ece.ubc.ca, hamarneh@sfu.ca}}


\maketitle
\begin{abstract}
Deep learning models have achieved great success in automating skin lesion diagnosis. However, the ethnic disparity in these models' predictions, where lesions on darker skin types are usually underrepresented and have lower diagnosis accuracy, receives little attention. In this paper, we propose \modelname{}, a disentanglement deep learning framework with contrastive learning that utilizes an additional network branch to remove sensitive attributes, i.e. skin-type information from representations for fairness and another contrastive branch to enhance feature extraction. We compare \modelname{} to three fairness methods, namely, resampling, reweighting, and attribute-aware, on two newly released skin lesion datasets with different skin types: Fitzpatrick17k and Diverse Dermatology Images (DDI). We adapt two fairness-based metrics $DPM$ and $EOM$ for our multiple classes and sensitive attributes task, highlighting the skin-type bias in skin lesion classification. Extensive experimental evaluation demonstrates the effectiveness of \modelname{}, with fairer and superior performance on skin lesion classification tasks.
\end{abstract}

\section{Introduction}
Cancer is the leading cause of death worldwide, accounting for nearly 10 million deaths in 2020, or about one in six deaths, and the skin is among the 6 most common organs invaded by cancer~\cite{whocancer,sung2021global}. Nevertheless, the survival rate of patients can be considerably increased by early detection and treatment of skin lesions~\cite{balch2009final}. Traditional diagnosis and detection of skin cancer have been carried out by dermatologists via manual screening and visual inspection, which is time-consuming, complex and error-prone. With the advancement in computer vision, tedious examination procedures may be avoided by an automatic diagnosis of identifying possibly cancerous lesions. End-to-end deep neural networks (DNNs) were developed to capture more powerful features and output predictions based directly on the input images~\cite{el2020deep,gessert2019skin}. Despite their success, DNNs are vulnerable to biases that render their decisions “unfair”. For example, a recent study has shown that patients with darker skin types (tones or colors) experience a significant drop-off in diagnosis accuracy compared to those of light types~\cite{daneshjou2021disparities}, which limits the successful clinical deployment of DNNs~\cite{groh2021evaluating,li2021estimating}.

Fairness is challenging to address for skin lesion classification, given the lack of annotated data with diverse skin types. Kinyanjui et al.~\cite{kinyanjui2020fairness} first proposed a pipeline for automatic estimation of skin-type labels based on the individual typology angle (ITA) for skin disease datasets. Other works further adopted the ITA to generate skin-type labels and proposed skin-type bias mitigating strategies~\cite{li2021estimating,chabi2022towards,bevan2022detecting}. However, they validated their approaches mainly on datasets with a small fraction of dark skin-type samples, limiting their conclusions' statistical power and generalizability. Furthermore, Groh et al.~\cite{Groh2022TowardsTI} found that ITA might be unreliable for annotating large-scale image datasets. Recently, Groh et al.~\cite{groh2021evaluating} and Daneshjou et al.~\cite{daneshjou2021disparities} proposed two new datasets with varied skin types and manually annotated Fitzpatrick skin-type labels 1 through 6, named Fitzpatrick17k and Diverse Dermatology Images (DDI). Some researchers~\cite{bevan2022detecting,wu2022fairprune} used these datasets to test the efficacy of their proposed "fairness" approaches. However, they simplified the experimental settings by only testing generalizability on a subset of the data, i.e. training models on intermediate skin types (merging types 3 and 4) and testing on dark skins (merging types 5 and 6) without using data of fairer skin types 1 and 2~\cite{bevan2022detecting} or splitting multiple skin types into two sensitive groups; i.e. binary sensitive attribute classification task~\cite{wu2022fairprune}.

\emph{Fairness through blindness} is a recent promising direction for mitigating unfairness in DNNs. Blindness to sensitive attributes is achieved by complementing the target DNN branch with a branch dedicated to classifying the sensitive attribute, e.g. gender. The framework captures semantic information to accurately perform the target task, e.g. classification, while simultaneously \textit{minimizing} the accuracy of the sensitive-attribute classifier, learning sensitive-attribute agnostic representations. This can be achieved through adversarial learning~\cite{wang2019balanced,elazar2018adversarial,beutel2017data} or disentangled representation learning~\cite{xu2020investigating,alvi2018turning,puyol2021fairness}. Nevertheless, as noted in Elazar et al.~\cite{elazar2018adversarial} and Wang et al. \cite{wang2020towards}, one of the main limitations of these methods is that they may hurt the representation learning of the target task. In particular, after sensitive attribute information has been removed from representations, the model might consider combinations of other attributes as a proxy and discard their information by mistake. This will result in accuracy deterioration when the information is related to the target task. For example, lesion color and visual features are critical for skin lesion classification~\cite{barata2018survey}, but are also useful for inferring the individual's skin type. To address this issue, we introduce contrastive learning that has shown impressive success in self-supervised pre-training tasks~\cite{he2020momentum} and many other fields~\cite{khosla2020supervised,thota2021contrastive,wang2021contrastive} for blindness-based fairness algorithms. Our contrastive loss encourages the representations of samples from the same target class to be proximate regardless of the value of their sensitive attributes, and those from different target classes to be distant. The network is thus enforced to retain discriminative semantic information about the target task.

In this work, we propose \modelname{}, a disentanglement framework with contrastive learning for fairness in dermatology, which not only discourages discrimination against sensitive attributes, i.e. skin-type information, but also preserves encoding the visual characteristics related to classification in the hidden representations through contrastive learning. The network contains a feature extractor and three branches: \emph{target branch}, \emph{sensitive attribute (SA) branch}, and \emph{contrastive branch}. Specifically, the feature extractor encodes the input images into representations. The target branch utilizes these representations to make skin condition predictions. To mitigate the skin-type unfairness, the SA branch enforces the feature extractor to discard the skin-type information by minimizing the likelihood that the model correctly predicts skin types based on the representations. The contrastive branch utilizes supervised contrastive learning to improve the quality of representations and boost classification accuracy. To investigate the fairness in skin type comprehensively and in-depth, we conduct experiments using in-domain and out-domain classification tasks on multiple skin conditions and skin types. In in-domain classification tasks, the training and test sets have the same skin types, whereas in out-domain, they have distinct skin types. We also adapt two fairness metrics from binary classification and modify them for non-binary classification. We additionally evaluate the performance of three widely used fairness methods (reweighting, resampling, and attribute-aware) to compare our \modelname{} against. 

Our contributions could be summarized:
    (1)
    We propose a novel framework \modelname{}, featuring disentangled representation learning and contrastive learning, to promote fairness and boost classification accuracy.
    (2)
    To the best of our knowledge, we are the first to examine unfairness in skin lesion datasets using a variety of approaches and extensive experiments.
    (3)
    We employ three fairness metrics to better compare models, including two that we adapted from the binary sensitive attributes task. 
    (4) 
    \modelname{} achieves the best classification accuracy and fairness scores compared to other fairness-based methods and the baseline. Our code is available at \url{https://github.com/siyi-wind/FairDisCo}.

\section{Related Works}
\subsection{Skin Lesion Diagnosis}

Skin cancer is mostly diagnosed clinically by experts, starting with a preliminary clinical screening and possibly followed by a dermoscopic evaluation, a biopsy, or histopathological examination~\cite{esteva2017dermatologist}. Traditionally, automatic image-based diagnosis is comprised of pre-processing, feature extraction, lesion segmentation, and classification. Dermatologists or machine learning classifiers make a diagnosis based on hand-crafted lesion features~\cite{jamil2014comparative}, following the ABCD rule~\cite{nachbar1994abcd}, the CASH algorithm~\cite{henning2007cash}, or the seven-point checklist~\cite{healsmith1994evaluation}.

Deep learning classification techniques, which do not require segmentation or hand-crafted feature extraction stages, are currently the most popular approaches to automate skin disease detection. Kawahara et al.~\cite{kawahara2016deep} applied a pre-trained convolutional neural network (CNN) as a feature extractor, which yields better results than prior works relying on general engineered features. Esteva et al.~\cite{esteva2017dermatologist} then trained an end-to-end CNN outperforming 21 board-certified dermatologists on biopsy-proven clinical images. However, its excellent performance demands a sizable training dataset. Harangi et al.~\cite{harangi2018skin} further fused the outputs of several CNNs to reach higher accuracy with limited skin lesion images. One of the state-of-the-art algorithms in this field is the work by Gessert et al.~\cite{gessert2020skin}. They used different cropping strategies for input images with different resolutions. The final optimal model is aggregated from eight different CNNs inputting diverse image sizes through an ensemble method. Despite automatic classification methods flourishing in the past several years, the fairness issue in these models has not received much attention~\cite{bhardwaj2021skin,adegun2021deep}.

\subsection{Fairness}

Fairness in machine learning is an increasing concern for the public, governments, and scientists~\cite{mehrabi2021survey}. Many works have been proposed to reduce unfairness in deep learning, which can be categorized into 3 groups: pre-processing~\cite{kamiran2012data,bellamy2019ai}, in-processing~\cite{wadsworth2018achieving,bendekgey2021scalable,alvi2018turning,xu2020investigating}, and post-processing~\cite{petersen2021post,wang2022fairness}. 

\textit{Pre-processing} methods aim to transform the data so that the underlying discrimination is removed. A representative work is from Kamiran et al.~\cite{kamiran2012data} who introduced and investigated four intuitive pre-processing methods to get classifiers in an optimal trade-off between accuracy and non-discrimination. 
To train a fairer model, \textit{In-processing} techniques either modified the model architecture or added fairness-related penalties. Wadsworth et al.~\cite{wadsworth2018achieving} added an adversarial architecture after a recidivism-predicting neural network, which helps to eliminate racial bias in the prediction. To get a sensitive-attribute agnostic representation, Sarhan et al.~\cite{sarhan2020fairness} disentangled meaningful and sensitive information by enforcing orthogonality constraints as a proxy for independence. Adding a fairness regularizer is a prospective direction. However, the main problem is that common fairness formulas are not differentiable, so they cannot be directly used in the objective function. Bendekgey et al.~\cite{bendekgey2021scalable} introduced three new surrogates of fairness constraints for non-convex models, which can be applied to challenging computer vision and natural language processing problems. As for the \textit{post-processing} methods, they aim to utilize the model’s outputs and sensitive attributes to calibrate the model’s prediction during inference. Petersen et al.~\cite{petersen2021post} cast the individual fairness post-processing problem as a graph smoothing problem corresponding to graph Laplacian regularization that preserves the desired “treat similar individuals similarly” interpretation.

While creating models that are fair to age, sex, or race has become increasingly common, skin-type fairness in skin lesion diagnosis draws little attention. Kinyanjui et al.~\cite{kinyanjui2020fairness} found no clear relationship between skin type and segmentation performance in datasets. Bevan et al.~\cite{bevan2022detecting} presented a modified variational autoencoder to uncover skin-type bias and executed a partial experiment, an out-domain classification on two skin-type groups. Wu et al.~\cite{wu2022fairprune} proposed FairPrune, a strategy that pruned parameters during training to reduce the accuracy gap between different skin types, and validated the model on binary sensitive attributes by grouping six-scale Fitzpatrick annotations to two groups (light and dark). In this paper, we directly use the skin-type annotations in the dataset to study multi-class classification with multiple sensitive attributes and perform in-domain and out-domain classification experiments on two datasets.

\section{Methodology}\label{methodology}
In a multi-class skin lesion classification ($M$ classes), the model is required to output a skin condition prediction $y$ based on an RGB skin image $\boldsymbol{X} \in \mathbb{R}^{H \times W \times 3}$. We treat the skin type as a sensitive attribute $s$, including $N$ groups with diverse types. Our goal is to model $p(y|\boldsymbol{X})$ without being affected by $s$. In Section~\ref{content: proposed model}, we outline our proposed framework \modelname{}, in which we incorporate three branches to disentangle skin-type information from the learned latent representations and enhance the feature extraction for higher classification accuracy. In Section~\ref{content: other algorithms}, we further explore the existence and effects of the unfair skin-type issue in skin disease datasets. Specifically, we conduct a systematic study by comparing the proposed framework against the baseline and three simple but widely used fairness approaches: two pre-processing approaches (reweighting and resampling) and one in-processing approach (attribute-aware).

\begin{figure}[t]
\centering
\includegraphics[width=1\linewidth]{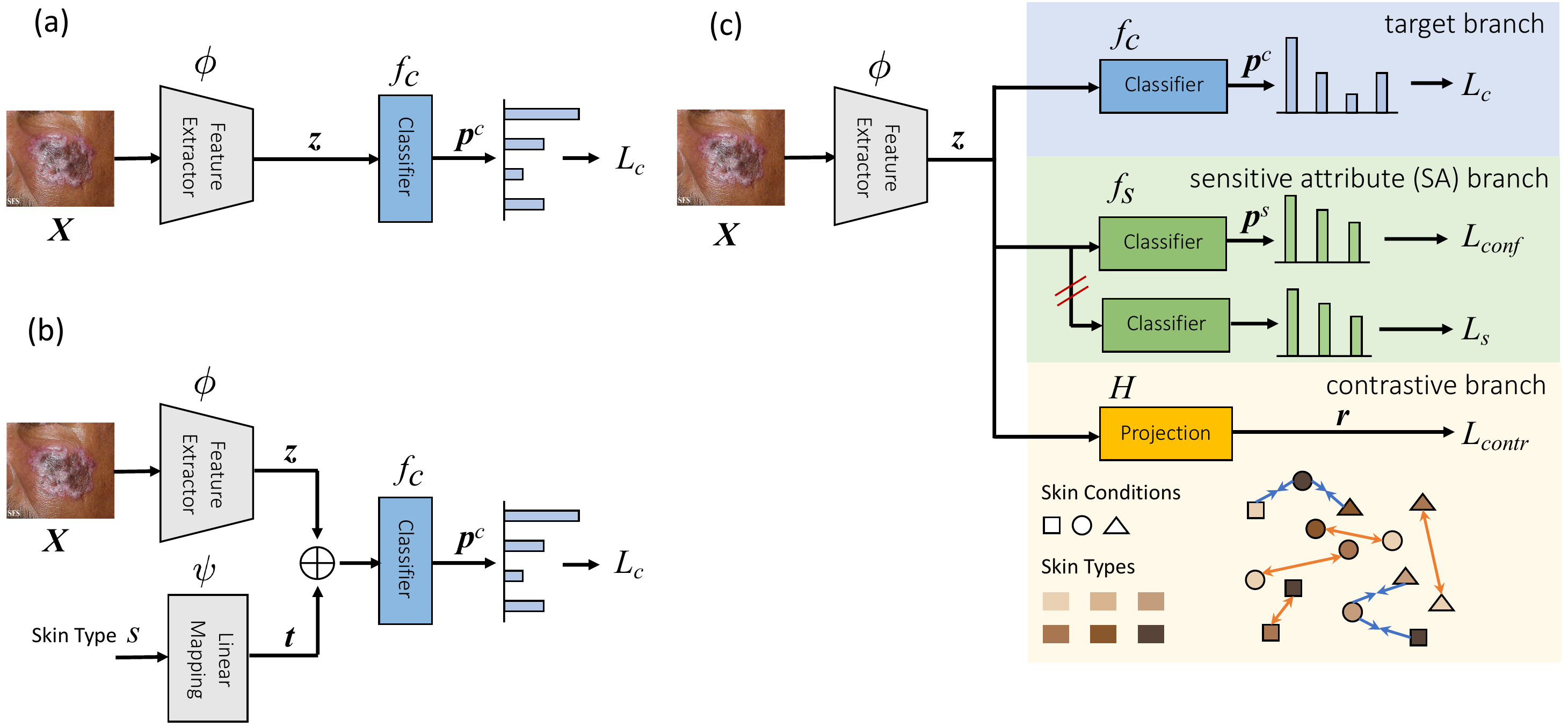}
\caption{Diagram of 3  skin disease classifiers: (a) Baseline; (b) Attribute-aware method; (c) Our proposed disentanglement network with contrastive learning (\modelname{}).}
\label{fig:model_architecture}
\end{figure}

In the baseline (BASE) method (Figure~\ref{fig:model_architecture}(a)), an image is input to a feature extractor $\phi$ to get a representation $\boldsymbol{z} = \phi(\boldsymbol{X})$ that is then passed to a classifier $f_c$, which includes one linear layer and a softmax activation function, to get a skin condition prediction $\boldsymbol{p}^c$. We utilize a cross-entropy loss $L_c$ on skin conditions to optimize the whole architecture. Figure~\ref{fig:model_architecture}(b) depicts an attribute-aware method (ATRB) with an additional skin-type  branch (further discussed in Section \ref{content: other algorithms}).

\subsection{Proposed \modelname{} model}\label{content: proposed model}
The intuition behind \modelname{} is to train a model to avoid capturing spurious correlations between skin conditions and types while learning semantic information from images. To achieve that, we incorporate three branches following a feature extractor (Figure~\ref{fig:model_architecture}(c)): the target branch to predict skin conditions as the baseline's classifier does, the SA branch to decouple skin-type information from representations, and the contrastive branch to enhance feature extraction. We will introduce the sensitive and the contrastive branches in detail.

\subsubsection{Disentangled Representation Learning:} The SA branch consists of a classifier $f_s$ to predict the skin type $\boldsymbol{p}^s$ based on the representation $\boldsymbol{z}$. Methods to obtain disentangled representation in \emph{fairness through blindness} could be grouped into two. One is adversarial learning that utilizes an adversarial loss such as cross-entropy for sensitive attributes on the SA branch and a gradient-reversal layer (GRL) between the feature extractor and the classifier, i.e. the feature extractor receives opposite gradients from the SA branch~\cite{wang2019balanced,elazar2018adversarial,beutel2017data}. When the adversarial loss is minimized, the classifier's ability to predict sensitive attributes is maximized, and the feature extractor's ability to attain sensitive attributes is minimized. Another direction is disentangled representation learning by using two different losses without GRL~\cite{xu2020investigating,alvi2018turning}. We follow the latter direction. \modelname{} minimizes a confusion loss given in Equation~\ref{equation: confusion loss} to confuse the feature extractor and remove the skin-type information from representations.

\begin{align}\label{equation: confusion loss}
    L_{conf}=-\sum_{i=1}^N \frac{1}{N} \log (\boldsymbol{p}^s_i). 
\end{align}
This loss is minimized when the classifier outputs equal probability $\boldsymbol{p}^s_i$ for all skin types $i$, i.e., the representations are free of skin-type information. Notice the classifier $f_s$ might learn a tricky solution like setting all the weights to zero, then $\boldsymbol{p}^s$ is a zero vector, and $L_{conf}$ becomes the smallest even though the representation still contains skin-type information. Thus, we add a skin-type predictive cross-entropy loss $L_s$ only optimizing $f_s$. These two losses are opposite and serve the same purpose as the adversarial loss and GRL in the first research direction.

\subsubsection{Contrastive Feature Extraction Enhancement:}
Although disentangled representation learning keeps the feature extractor blind to skin-type information, it might hurt the extractor to encode semantic information by discarding important features related to the target task that could indicate skin type. Thus we add a contrastive loss~\cite{khosla2020supervised} that promotes intra-class cohesion and inter-class diversity to protect target features and improve the representation learning. In the contrastive branch (Figure~\ref{content: proposed model}(c)), we first project representations from the feature extractor into a low dimensional latent space $\boldsymbol{r} = H(\boldsymbol{z})$. For each embedding in the mini-batch, we split other embeddings with the same disease labels into a positive set $P_y$ and the rest into a negative set $N_y$ (regardless of the skin type). We calculate a contrastive loss as follows.
\begin{align}
    L_{contr}=-\frac{1}{|P_y|} \sum_{\boldsymbol{p} \in P_y}{\log\frac{\exp(\Psi(\boldsymbol{r}, \boldsymbol{p})/\tau)}{\exp(\Psi(\boldsymbol{r}, \boldsymbol{p})/\tau)+\sum_{\boldsymbol{n} \in N_y}{\exp(\Psi(\boldsymbol{r}, \boldsymbol{n})/\tau)}}}.
\end{align}

\noindent $\Psi(\cdot, \cdot)$ is the cosine similarity between two vectors. $\tau>0$ is a temperature parameter. We minimize the contrastive loss to optimize the feature extractor and the projection head, thus enforcing the network to keep as many factors to push samples in the same class together (orange lines in Figure~\ref{fig:model_architecture}(c)) and push away those in other classes without considering differing skin types (blue lines in Figure~\ref{fig:model_architecture}). Our final loss function for \modelname{} is:
\begin{align}\label{equation: total}
    L_{total}=L_c(\theta_\phi,\theta_{f_c})+\alpha L_{conf}(\theta_\phi,\theta_{f_s})+L_s(\theta_{f_s})+\beta L_{contr}(\theta_\phi,\theta_H).
\end{align}
We use $\alpha$ and $\beta$ to adjust contributions of confusion loss and contrastive loss. Notice $L_s$ is only used to optimize $f_s$.

\subsection{An investigation for three approaches}\label{content: other algorithms}
To better understand the unfairness issue in skin lesion datasets and enrich our assessment of different fairness models, we study three widely used pre-processing and in-processing fairness algorithms. 

\emph{Resampling Algorithm (RESM)} considers samples in the same skin type and condition are in the same group and then oversamples minorities and undersamples majorities to construct a balanced dataset~\cite{kamiran2012data}, forcing the model to treat groups equally. We set sampling weights for groups as their inverse frequency.

\emph{Reweighting Algorithm (REWT)} seeks to make skin types and conditions independent to prevent the model from learning discriminatory features~\cite{kamiran2012data}. The expected probability of one group is a multiplication between the probabilities of the skin condition group and the skin-type group. However, we usually observed a lower probability due to the correlation between two attributes. To cut off the connection between them, we assign each sample the weight as follows:
\begin{align}\label{equation: reweighting-weight}
    w(X) = \frac{P_{\exp}(X(s)=s_i, X(y)=y_j)}{P_{obs}(X(s)=s_i, X(y)=y_j)}.
\end{align}
\noindent The weight will be multiplied by the cross-entropy loss $L_c$ to adjust the loss contribution for each group.  

\emph{Attribute-aware Algorithm (ATRB)} was described in~\cite{xu2020investigating} where it adds skin-type information to the model so that its prediction will not be dominated by major groups. In Figure \ref{fig:model_architecture}(b), the feature extractor $\phi$ outputs a representation vector $\boldsymbol{z}$. The skin type $s$ is first converted to a one-hot vector and then mapped through one linear layer $\psi$ to a representation vector $\boldsymbol{t}$ that is the same size as $\boldsymbol{z}$. These two representations are then summed together to get a sensitive attribute-aware representation, which is finally sent into a classifier to get a skin condition prediction. The model is trained through $L_c$ as the baseline.

\section{Experiments}
\subsubsection{Datasets:}
We study two skin lesion datasets: Fitzpatrick17k dataset~\cite{groh2021evaluating} and Diverse Dermatology Images (DDI) dataset~\cite{daneshjou2021disparities}. In the Fitzpatrick17k dataset, Groh et al.~\cite{groh2021evaluating} compiled 16,577 clinical images with skin condition labels and annotated them with Fitzpatrick skin-type labels. There are 114 different skin conditions, and each one has at least 53 images. They further divided these skin conditions into two more advanced categories: 3 (malignant, non-neoplastic, benign) and 9. Fitzpatrick labelling system is a six-point scale initially developed for classifying sun reactivity of skin and adjusting clinical treatment according to skin phenotype~\cite{fitzpatrick1988validity}. Recently, it has been used in computer vision to evaluate algorithmic fairness \cite{hazirbas2021casual}. The samples in the dataset are labelled by 6 Fitzpatrick skin types and 1 unknown type. In our experiments, we ignore all samples of the unknown skin type. The DDI dataset contains 656 images with diverse skin types and pathologically confirmed skin condition labels, including 78 detailed disease labels and malignant identification. They grouped 6 Fitzpatrick scales into 3 groups: Fitzpatrick-12, Fitzpatrick-34, and Fitzpatrick-56, where each contains a pair of skin-type classes, i.e., \{1,2\}, \{3,4\} and \{5,6\}, respectively.

\subsubsection{Metrics:} We use accuracy to measure models' skin condition classification performance. To quantify fairness, we adapt earlier fairness metrics, which were restricted to binary classification or binary sensitive attributes~\cite{du2020fairness}, to our task, i.e., $>$2 disease classes and $>$2 skin types, resulting in 3 metrics: (i) Predictive Quality Disparity ($PQD$) measures the prediction quality difference between each sensitive group, which we compute as the ratio between the lowest accuracy to the highest accuracy across different skin-type groups, i.e.,
\begin{align}\label{equation: PQD}
    PQD=\frac{\min(acc_j, j \in S)}{\max(acc_j, j \in S)},
\end{align}
\noindent where $S$ is the set of skin types. 
(ii) Demographic Disparity (DP) computes the percentage diversities of positive outcomes for each sensitive group.
(iii) Equality of Opportunity (EO) asserts that different sensitive groups should have similar true positive rates. We calculate $DPM$ and $EOM$ across multiple skin conditions, $m\in
\{1,2,\cdots,M\}$, as follows:
\begin{align}\label{equation: DP_m}
    DPM=\frac{1}{M}\sum_{i=1}^{M}{\frac{\min[p(\hat{y}=i|s=j), j \in S]}{\max[p(\hat{y}=i|s=j), j \in S]}} \\
    EOM = \frac{1}{M}\sum_{i=1}^{M}{\frac{\min[p(\hat{y}=i|y=i,s=j), j \in S]}{\max[p(\hat{y}=i|y=i,s=j), j \in S]}},
\end{align}
\noindent where $y$ is the ground-truth skin condition label and $\hat{y}$ is the model prediction. A model is fairer if it has higher values for the above three metrics.

\subsubsection{Implementation Details:}
For the Fitzpatrick17k dataset, we carry out a three-class classification and follow Groh et al.~\cite{groh2021evaluating} in performing two experimental tasks. The first is an in-domain classification, in which the train and test sets are randomly split in an 8:2 ratio (13261:3318). The second is an out-domain classification, where we train models on samples of two skin types and test on samples of other skin types. We only use samples from those skin conditions present in both the train and the test sets and finally conduct three experimental settings $\mathcal{A}-\mathcal{C}$. $\mathcal{A}$ trains models on skin type 1-2 (7,755 samples) and tests on others (8,257 samples), $\mathcal{B}$ trains models on skin type 3-4 (6,089 samples) and tests on others (10,488 samples), and $\mathcal{C}$ trains models on skin type 5-6 (2,168 samples) and tests on others (14,409 samples). For the DDI dataset, we perform an in-domain binary classification (malignant vs non-malignant) using the same train-test ratio of 8:2 (524:132). For all the models, we use a pre-trained ResNet-18 without the final fully connected layer as the feature extractor. The dimension of the representation vector is 512. A Multi-Layer Perceptron (MLP) with a single hidden layer of size 512 and an output layer of size 128 serves as the projection head of the contrastive branch in \modelname{}. We follow Groh et al.~\cite{groh2021evaluating} in using a weighted random sampler to ensure equal numbers of samples for each skin condition in one mini-batch, except in RESM. Images are augmented through random cropping, rotation, and flipping to boost data diversity, then resized to $224 \times 224 \times 3$. We use Adam~\cite{kingma2014adam} optimizer to train the model with an initial learning rate $1 \times 10^{-4}$, which changes through a linear decay scheduler whose step size is 2, and decay factor $\gamma$ is 0.9. We deploy models on a single TITAN V GPU and train them with a batch size of 64. We set the training epochs for the Fitzpatrick17k dataset to 20 and the DDI dataset to 15.

\subsection{Results on the Fitzpatrick17k dataset}\label{content: in-domain fitz}

\subsubsection{Data Statistic:} In Table~\ref{table:distributioin}, numerous data biases occur in skin type and skin condition. The non-neoplastic group dominates the skin conditions with a 73\% share, and the type-2 group has the most samples of all skin-type groups, with a proportion that is 7.5 times higher than that of the minor type-6 group. Additionally, there are apparent discriminations between target and sensitive attributes. For instance, the group whose condition is malignant and type is 1 has an expected probability of 0.134 yet its observed probability is lower, 0.128, indicating that these two attributes are relevant (calculated as the way in REWT).

\setlength{\tabcolsep}{1pt}
\begin{table}[t]
\begin{center}
\caption{Data distributions for skin type and skin condition from the Fitzpatrick17k (Fitz) and DDI datasets.}
\label{table:distributioin}
\begin{tabular*}{\hsize}{@{}@{\extracolsep{\fill}}|l|c|c|c|c|c|c|c|c|@{}}
\hline
\multirow{2}{*}{~} & \multirow{2}{*}{Skin Condition} & \multicolumn{7}{c|}{Skin Type}\\
\cline{3-9}
 ~ & ~ & T1 & T2 & T3 & T4 & T5 & T6 & Total \\
\hline
\multirow{4}{*}{Fitz} & Benign & 444 & 671 & 475 & 367 & 159 & 44 & 2160\\
~ & Malignant & 453 & 742 & 456 & 301 & 147 & 61 & 2160\\
~ & Non-neoplastic & 2050 & 3395 & 2377 & 2113 & 1227 & 530 & 11692\\
~ & Total & 2947 & 4808 & 3308 & 2781 & 1533 & 635 & 16012\\
\hline
\multirow{4}{*}{DDI} 
& ~ & \multicolumn{2}{c|}{T12} & \multicolumn{2}{c|}{T34} & \multicolumn{2}{c|}{T45} & Total \\
\cline{3-9}
~ & Malignant & \multicolumn{2}{c|}{49} & \multicolumn{2}{c|}{74} & \multicolumn{2}{c|}{48} & 171\\
~ & Non-malignant & \multicolumn{2}{c|}{159} & \multicolumn{2}{c|}{167} & \multicolumn{2}{c|}{159} & 485\\
~ & Total & \multicolumn{2}{c|}{208} & \multicolumn{2}{c|}{241} & \multicolumn{2}{c|}{207} & 656\\
\hline
\end{tabular*}
\end{center}
\end{table}
\setlength{\tabcolsep}{1pt}

\setlength{\tabcolsep}{1pt}
\begin{table}[t]
\begin{center}
\caption{In-domain classification result comparison on the Fitzpatrick17k dataset. Accuracy, $PQD$, $DPM$, and $EOM$ are expressed in percentages (\%).}
\label{table:normal_fitz}
\begin{tabular*}{\hsize}{@{}@{\extracolsep{\fill}}|l|c|c|c|c|c|c|c|c|c|c|@{}}
\hline
\multirow{2}{*}{Model} & \multicolumn{7}{c|}{Accuracy (\%) $\uparrow$} & $PQD$ & $DPM$ & $EOM$\\
\cline{2-8}
 ~ & Avg & T1 & T2 & T3 & T4 & T5 & T6 & $\uparrow$ & $\uparrow$ & $\uparrow$ \\
\hline
GROH~\cite{groh2021evaluating} & 60.07 & 56.19 & 60.40 & 57.58	& 61.51 & 65.72 & 70.83 & 79.32 & 60.75 & 77.86\\
\hline
BASE & 85.11 & 82.94 & 82.21 & 86.45 & 87.77 & 89.75 & 88.33 & 91.60 & 46.99 & 67.88 \\
RESM & 85.45 & 82.27 & 82.63 & 86.45 & 89.39 & 90.11 & \underline{89.17} & 91.31 & 46.98 & 67.56  \\
REWT & 85.23 & 81.77 & 83.04 & 85.57 & 89.39 & 89.40 & \underline{89.17} & 91.47 & 48.92 & 70.64 \\
ATRB & 85.30	& 82.78	& 81.70	& 85.86	& \underline{90.65} & \underline{90.81} & 85.83 & 90.00 & 56.67 & 67.91 \\
\modelwocl{} & 85.61	& 83.44 & \underline{83.17}	& 85.42	& 89.57	& \underline{90.81} & 86.67 & 91.60 & \underline{57.64} & 74.52  \\
\modelname & \underline{85.76} & \underline{84.45} & 82.94 & \underline{86.60} & 88.49 & 90.11 & 87.50 & \underline{92.04} & 55.53 & \underline{75.62} \\
\hline
\end{tabular*}
\end{center}
\end{table}
\setlength{\tabcolsep}{1pt}

\subsubsection{In-domain Classification Results and Discussion:} 
\begin{figure}[t]
\centering
\includegraphics[width=1 \linewidth]{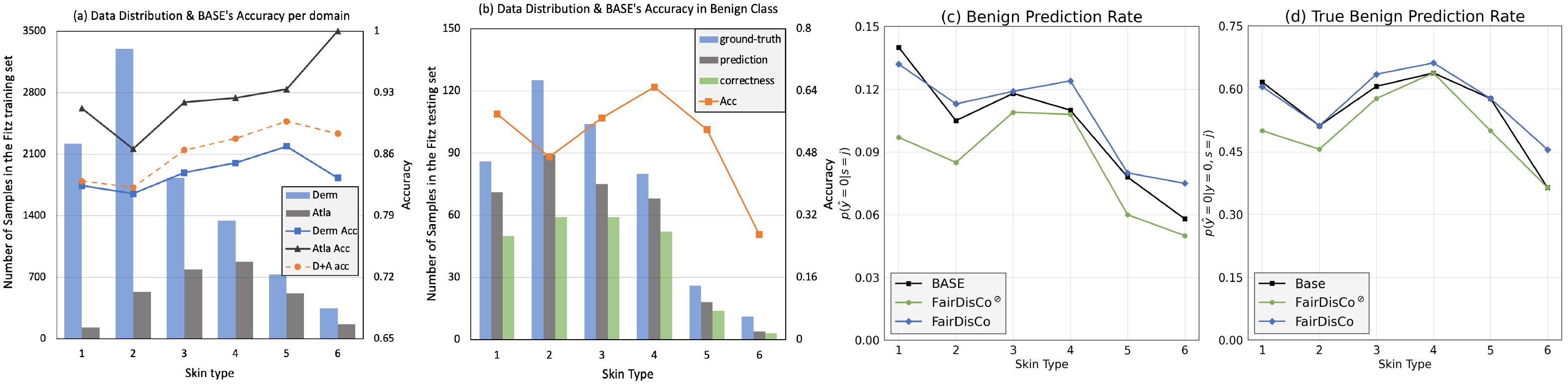}
\caption{(a) Training data distributions in Derm and Atla and corresponding accuracy of BASE; (b) Ground-truth, prediction, correctness, and accuracy distributions of the benign class in the Fitzpatrick17k test set; (c) Comparison of BASE, \modelwocl{}, and \modelname{} in benign prediction rate (the main term for calculating $DPM$) across skin types; (d) Comparison of BASE, \modelwocl{}, and \modelname{} in true benign rate (the main term for calculating $EOM$) across skin types.}
\label{fig:normal_analysis}
\end{figure}

We use accuracy and fairness metrics to examine the models in Section~\ref{methodology} and the model in~\cite{groh2021evaluating} (we name it GROH) that contains a fixed VGG-16 and one learnable linear classifier. \modelwocl{} is the framework \modelname~without the contrastive loss. Examining the results (Table~\ref{table:normal_fitz} and Figure~\ref{fig:normal_analysis}), we make the following observations.

(1) In Table~\ref{table:normal_fitz}, our baseline (BASE) exhibits inconsistent performance (accuracy) that varies across skin types, showing the existence of skin-type unfairness. Nevertheless, by changing the backbone and adjusting the training procedure, the baseline achieves much higher accuracy than GROH (on average 85.11\% vs 60.07\%). Since the average accuracy for GROH is too low compared with other methods, we exclude it from subsequent analysis.

(2) We observe the phenomena mentioned in~\cite{groh2021evaluating} where the accuracy of the major skin-type group is lower than the underrepresented skin-type group. The ambiguous systematic pattern in accuracy across skin types could be seen in all the models in Table~\ref{table:normal_fitz}, e.g. type 6 has higher accuracy than type 2. We attribute it to the domain distribution diversity, different diagnosis difficulties across skin types, and the possibly untrustworthy results for those groups with few test samples. We analyze the data in Figure~\ref{fig:normal_analysis}(a, b). 

The Fitzpatrick17k dataset is collected from two domains: DermaAmin (Der-m) and Atlas Dermatologico (Atla)~\cite{derm_dataset,atla_dataset}. As illustrated in Figure \ref{fig:normal_analysis}(a), Derm has the most samples in type 2, the fewest samples in type 6, and an average accuracy of 83.05\%, whereas Atla has the most samples in type 4, the fewest samples in type 1, and an average accuracy of 91.75\%. When we look into accuracy by skin type, type 6 in Atla (black line) has the highest accuracy compared to other skin types, but type 6 in Derm (blue line) is in fourth place. When combining these two domains, its total accuracy (orange line) is in second place, different from that in Derm or Atla. The diversities of distribution and classification difficulty across domains conceal the real pattern when we directly train models on two domains and only report the overall accuracy. In addition, Figure~\ref{fig:normal_analysis}(b) shows that BASE's prediction (gray bars) and ground-truth (blue bars) in benign class have similar distributions, with the most number of samples in type 2 and the least in type 6, indicating the model's prediction is affected by skin type. However, the correctness distribution is quite different. Type 2-4 have a similar number of correctness (green bars) despite type 2 having more benign predictions. Thus type 2 is less accurate than other skin types, i.e. it has more ground-truth and predictions, but less correctness. Then we could hypothesize the images of type 2 might be more challenging to classify than those of type 4 and 5, confusing the relationship between accuracy and statistical data distribution. Moreover, the small group size might make the test results unreliable. In our test case, the group whose skin condition is benign and skin type is 6 only has 11 samples, and the group whose condition is malignant and type is 6 just has 7 samples. A correct or wrong prediction will significantly impact their accuracy, making the results of underrepresented groups untrustworthy. 

(3) Compared with BASE, all fairness techniques (RESM, REWT, ATRB, \modelwocl{}, and \modelname) achieve higher average accuracy (Table~\ref{fig:normal_analysis}). Additionally, these methods, except RESM, surpass the baseline on more than two fairness metrics. RESM performs worse than BASE on three fairness metrics due to the serious data imbalance, as illustrated in Table~\ref{table:distributioin}.

(4) In Table~\ref{fig:normal_analysis}, \modelname{} achieves the highest accuracy (outperforms BASE by 0.65\% on average accuracy) and fairness (outperforms BASE by 0.44\%, 8.54\%, and 7.74\% on three fairness metrics). The difference between $PQD$ is much smaller than $DPM$ and $EOM$ because the latter two give equal weight to each class when summing them together; thus, they will obviously decrease when one class is extremely unfair. In addition, \modelname~outperforms \modelwocl{} on average accuracy and two fairness metrics, demonstrating the effectiveness of contrastive learning. The gain in fairness might be because the contrastive loss only uses target label information to cluster the data, resulting in representations invariant across skin types when regarding each skin-type group as a domain.

(5) We further visualize the benign prediction rate $p(\hat{y}=0|s=j), j \in S$ used to calculate $DPM$ and true benign rate $p(\hat{y}=0|y=0, s=j)$ used to calculate $EOM$ in Figure~\ref{fig:normal_analysis}(c, d). BASE has diverse values across skin types, indicating the model is affected by skin-type information. Unlike BASE (black line), the benign prediction rates for both \modelname~(blue line) and \modelwocl{} (green line) present little variation across skin types. In particular, \modelname~boosts the minor skin-type group's benign prediction rate and true benign rate and performs the best in decreasing the model's diversities across skin types, which exemplifies that our model reduces skin-type unfairness. Finally, we could conclude that skin type in the Fitzpatrick17k dataset influences the models' decisions despite no systematic pattern between accuracy and sensitive group size.

\subsubsection{Out-domain Classification Results and Discussion}
We conduct experiments on the baseline and other four fairness methods and show their accuracies and fairness metrics. From Table~\ref{table:general_fitz}, we could acquire the following findings. 

\setlength{\tabcolsep}{4pt}
\begin{table}[t]
\begin{center}
\caption{Out-domain classification result comparison on the Fitzpatrick17k dataset. The first column (E) presents the experiment index. FairDisCo is abbreviated as FDC.}
\label{table:general_fitz}
\begin{tabular*}{\hsize}{@{}@{\extracolsep{\fill}}|l|l|c|c|c|c|c|c|c|c|c|c|@{}}
\hline
\multirow{2}{*}{E} & \multirow{2}{*}{Model} & \multicolumn{7}{c|}{Accuracy (\%) $\uparrow$} & $\scriptstyle PQD$ & $\scriptstyle DPM$ & $\scriptstyle EOM$\\
\cline{3-9}
 ~ & ~ & Avg & T1 & T2 & T3 & T4 & T5 & T6 & $\uparrow$ & $\uparrow$ & $\uparrow$ \\
\hline
\multirow{6}{*}{$\mathcal{A}$} & BASE &  80.33 &	- &	- &	80.20 &	80.65 &	\underline{79.58} &	\underline{81.42} &	97.74 & 67.48	& 69.09 \\
~ & RESM & 79.13 &	- &	- &	80.44 &	79.11 &	77.76 &	75.75 &	94.17 & 79.49 & 63.29  \\
~ & REWT &  79.16 & - & - & 79.05 & 79.04 & 79.19	& 80.16 &	\underline{98.60} & 69.56 &	\underline{72.30} \\
~ & ATRB & 79.52 &	- &	- &	80.17 &	80.04 &	77.56 &	78.58 &	96.74 & 65.76 &	62.60 \\
~ & FDC$^{\oslash}$ &  \underline{80.50} &	- &	- &	80.93 &	80.80 &	79.00 &	80.79 &	97.80 & 67.34 & 71.15  \\
~ & FDC & 80.37 & - & - & \underline{81.41} & \underline{81.12} & 78.02	& 77.32 & 94.98 & \underline{74.34} & 62.11  \\
\hline
\hline
\multirow{6}{*}{$\mathcal{B}$} & BASE &  78.38 &	73.91 &	77.83 &	- &	- &	86.82 &	82.99 &	85.12 & 	61.33 & 71.51 \\
~ & RESM &  78.74 & 74.69 & \underline{78.62} & - & - & 85.91	& 81.10 &	86.94 & 71.03 &	\underline{74.96} \\
~ & REWT & 77.60 &	72.89 &	77.18 &	- & - &	85.58 &	83.31 &	85.17 & 53.97 &	64.90 \\
~ & ATRB &  78.23 & 72.99 & 77.79 & - & - & \underline{86.95} &	\underline{84.88} &	83.94 & 57.34 &	73.53\\
~ & FDC$^{\oslash}$ &  \underline{78.81} &	\underline{75.20} &	78.12 &	- & - &	85.52 &	84.57 &	\underline{87.93} & 53.80 & 70.48 \\
~ & FDC & 78.61 & 74.62 & 78.56 & - & - & 85.52 &	80.79 & 87.25 & \underline{71.44} & 69.53  \\
\hline
\hline
\multirow{6}{*}{$\mathcal{C}$} & BASE &  73.49 & 69.11 & 70.98	& \underline{74.65}	& 81.01	& - & -	& 85.31 & 61.78	& 79.33 \\
~ & RESM &  73.70 & 69.05 & 71.83 & 74.56 & 80.76 & - & -	& 85.50 & 63.23	& 77.65\\
~ & REWT &  69.75 & 61.70 & 66.95 & 72.30 & 79.93 & - & -	& 77.20 & 56.24	& 75.50\\
~ & ATRB &  72.92 & 68.18 & 70.90 & 73.49 & 80.65 & - & -	& 84.54 & 67.80	& 77.79\\
~ & FDC$^{\oslash}$ &  \underline{74.11} &	70.36 &	\underline{72.08} &	74.10 &	\underline{81.56} &	- & - &	86.27 &	67.18 &	79.02 \\
~ & FDC &  73.64	& \underline{70.63}	& 71.64	& 73.61	& 80.25	& - & -	& \underline{88.01}	& \underline{70.69}	& \underline{83.69} \\
\hline
\end{tabular*}
\end{center}
\end{table}
\setlength{\tabcolsep}{1.4pt}

(1) BASE has higher accuracy on those skin types similar to those it is trained on, such as type 4's accuracy in Experiment $\mathcal{C}$ is 11.90\% higher than type 1's. This reveals that the model's decision-making is relevant to the skin type, which is consistent with the phenomenon indicated in~\cite{groh2021evaluating}.

(2) For fairness approaches, ATRB performs worse than BASE, since it depends on skin-type information to pay equal attention to each group but cannot access skin type in the test set during training. RESM improves the accuracy and at least two fairness metrics in Experiments $\mathcal{C}$ and $\mathcal{D}$.

(3) \modelwocl{} reaches the best results (average accuracy and three fairness metrics) in three experiments. Its increase in minority accuracy serves as further evidence of the unfair skin-type problem. \modelname~is the second-best among all the algorithms, which raises the average accuracy on three experiments and has better fairness performance on $\mathcal{B}$ and $\mathcal{C}$ in comparison to BASE. We attribute its inferior performance compared with \modelwocl{} to that the contrastive loss only helps the model learn better representations of skin-type groups appearing in the training set. However, the out-domain classification task has a large diversity across skin types between the training and test sets.

\setlength{\tabcolsep}{4pt}
\begin{table}[t]
\begin{center}
\caption{In-domain classification result comparison on the DDI dataset.}
\label{table:normal_ddi}
\begin{tabular*}{\hsize}{@{}@{\extracolsep{\fill}}|l|c|c|c|c|c|c|c|c|@{}}
\hline
\multirow{2}{*}{Model} & \multicolumn{4}{c|}{Accuracy (\%) $\uparrow$} & $PQD$ & $DPM$ & $EOM$ & AUC\\
\cline{2-5}
 ~ & Avg & T12 & T34 & T56 & $\uparrow$ & $\uparrow$ & $\uparrow$ & $\uparrow$\\
\hline
BASE & 82.58 &	\underline{83.78} &	\underline{84.31}	& 79.55 & 94.34 &	79.28 &	80.35 & 0.81\\
RESM & 82.58 & \underline{83.78} &	80.39 &	84.09 & 95.60 &	79.28 &	\underline{93.01} & 0.80\\
REWT &  82.58 & 81.08 & 78.43	& \underline{88.64} & 88.49 & 83.39 & 82.45 & 0.82\\
ATRB & 81.82 & 75.68 & 82.35 & 86.36 & 87.62 & \underline{85.16} & 72.42 & 0.82\\
\modelwocl{} & 82.58	& 81.08	& 80.39	& 86.37 & 93.09	& 77.45	& 90.54	& 0.82 \\
\modelname & \underline{83.33} & \underline{83.78} & \underline{84.31} & 81.82 & \underline{97.04}	& 83.62 &	83.46 & \underline{0.85}\\
\hline
\end{tabular*}
\end{center}
\end{table}
\setlength{\tabcolsep}{1.4pt}

\subsection{Results on the DDI dataset}
\subsubsection{Data Statistic:} Compared with the Fitzpatrick17k dataset, the DDI dataset contains similar skin-condition data imbalance but less skin-type bias (Table~\ref{table:distributioin}). 74\% of cases are non-malignant, while the rates for the three skin-type groups are 1:1.16:1. Skin condition and skin type are also dependent in this dataset, e.g. the group whose condition is non-malignant and type is 12 has an expected probability of 0.083 whereas an observed probability of 0.075.

\subsubsection{In-domain Classification Results and Discussion:} The compared models are the same as those of the in-domain classification in Section~\ref{content: in-domain fitz} except GROH. We further add another classification performance measurement, area under the ROC curve (AUC). From Table~\ref{content: in-domain fitz}, we observe the following.

(1) For BASE, the major skin-type group (T34) is more accurate than the minor groups (T12 and T34), showing it ignores the underrepresented groups. 

(2) All the fairness approaches boost the minor group's (T56) accuracy. RESM and REWT improve two fairness metrics while maintaining the average accuracy, but ATRB does not outperform BASE.

(3) \modelname~not only yields the best classification performance (outperforms BASE by 0.75\% on average accuracy and 4\% on AUC) but also achieves the best fairness results (outperforms BASE by 2.7\%, 4.34\%, and 3.11\% on three fairness metrics). Compared to \modelwocl{}, the superior average accuracy, $PQD$, $DPM$, and AUC of \modelname{} further demonstrate that the contrastive loss plays an integral role in learning a better semantic representation.

\subsection{Loss Analysis}
\begin{figure}[t]
\centering
\includegraphics[width=1 \linewidth]{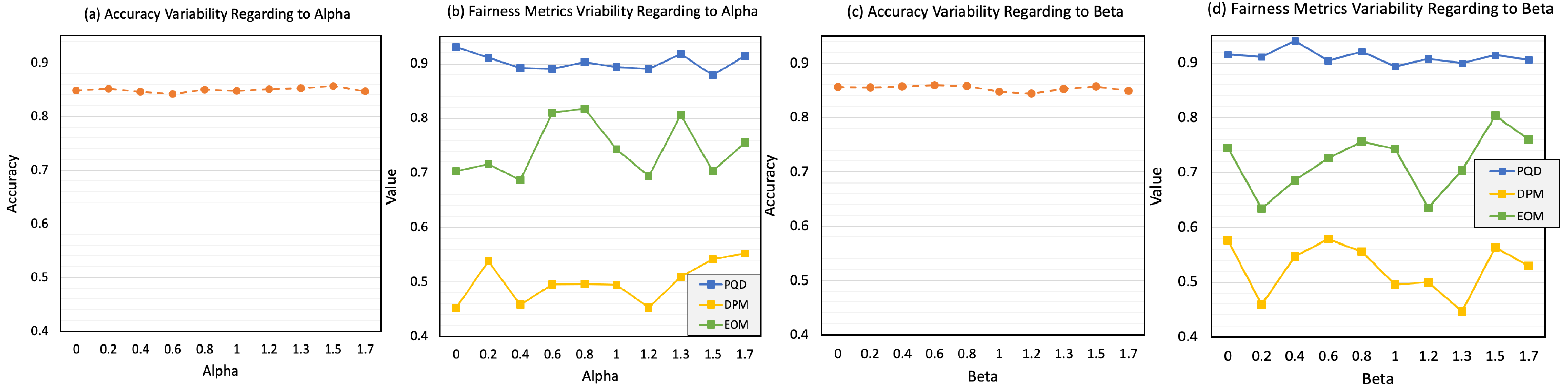}
\caption{(a) and (b) are average accuracy and fairness metrics trends based on confusion loss weight $\alpha$. (c) and (d) are the trends based on contrastive loss weight $\beta$.}
\label{fig:sensitive_analysis}
\end{figure}

To analyze the sensitivity of \modelname{} to the confusion loss weight $\alpha$ and contrastive loss weight $\beta$ of Equation~\ref{equation: total}, 
we run experiments on the Fitzpatrick17k dataset where we fix  $\alpha$ to $1.0$ while adjusting  $\beta$, and vice versa, and report the accuracy and fairness
metric values in Figure~\ref{fig:sensitive_analysis}. 
When $\alpha$ and $\beta$ change, 
even as the accuracy, $PQD$, $DPM$, and $EOM$ values vary by 1.84\%, 5.19\%, 13.1\% and 18.2\%, respectively, \modelname{} mostly continues to outperform BASE on the 4 metrics, and outperforms RESM, REWT, and ATRB
on at least two fairness metrics. We found no obvious relation between the model's performance and the loss composition. Though, unsurprisingly, excessively increasing the contribution of either loss can lead to a decreased model performance.

\section{Conclusion}
We investigated skin-type fairness in skin lesion classification datasets and models. We improved the fairness and boosted the classification accuracy by proposing a novel deep neural network, \modelname{}. \modelname{} leverages a network branch to disentangle skin-type information from the latent representation and a contrastive branch to avoid accuracy deterioration. To demonstrate the presence of unfairness in datasets and the effectiveness of \modelname{}, we compared the baseline and \modelname{} to 3 widely used fairness techniques (reweighting, resampling, and attribute-aware). Moreover, as there are no standard fairness metrics for evaluating multi-class classification with multiple sensitive attributes, and as using only the ratio of accuracy per skin type ignores the accuracy per skin condition, we proposed three fairness metrics $PQD$, $DPM$, and $EOM$ for our task which could also be applied to other fairness tasks. We compared models on in-domain and out-domain classification experiments using the Fitzpatrick17k and DDI, two datasets with diverse skin types and diseases, and clarified the relationship between accuracy and group size in the Fitzpatrick17k in-domain classification. The results indicate that the datasets have skin-type bias, and fairness methods could diminish the discrimination. Our proposed model \modelname{} achieved the best overall accuracy and fairness on in-domain classification tasks and surpassed the baseline on the out-domain classification task.

\clearpage
%
%
\bibliographystyle{splncs04}
\bibliography{egbib}
\end{document}